\documentclass{main_style}
\usepackage{hyperref}
\usepackage{graphicx}
\usepackage{amsmath}
\usepackage{float}
\usepackage{amssymb}
\usepackage{pifont}
\newcommand{\xmark}{\ding{55}}%
\usepackage[bottom]{footmisc}
\finalcopy 
\def \etal{\emph{et~al.}}
\begin{document}
\title{Sparse and noisy LiDAR completion with RGB guidance and uncertainty}

\author{Wouter Van Gansbeke \hspace{20pt} Davy Neven \hspace{20pt} Bert De Brabandere \hspace{20pt} Luc Van Gool\\
ESAT-PSI, KU Leuven\\
\small{firstname.lastname@esat.kuleuven.be}
}
\maketitle

\section*{\centering Abstract}
\textit{
 This work proposes a new method to accurately complete sparse LiDAR maps guided by RGB images. For autonomous vehicles and robotics the use of LiDAR is indispensable in order to achieve precise depth predictions. A multitude of applications depend on the awareness of their surroundings, and use depth cues to reason and react accordingly. On the one hand, monocular depth prediction methods fail to generate absolute and precise depth maps. On the other hand, stereoscopic approaches are still significantly outperformed by LiDAR based approaches. The goal of the depth completion task is to generate dense depth predictions from sparse and irregular point clouds which are mapped to a 2D plane. We propose a new framework which extracts both global and local information in order to produce proper depth maps. We argue that simple depth completion does not require a deep network. However, we additionally propose a fusion method with RGB guidance from a monocular camera in order to leverage object information and to correct mistakes in the sparse input. This improves the accuracy significantly. Moreover, confidence masks are exploited in order to take into account the uncertainty in the depth predictions from each modality. This fusion method outperforms the state-of-the-art and ranks first on the KITTI depth completion benchmark~\cite{kitti}. Our code with visualizations is available at \url{https://github.com/wvangansbeke/Sparse-Depth-Completion}.
}

\section{Introduction}
 Depth completion is predicting dense depth maps from a sparse point cloud. In many computer vision applications, precise depth values are of crucial importance. In recent years this task has gained attention due to industrial demand. Other computer vision tasks, among which 3D object detection and tracking, 2D or 3D semantic segmentation and SLAM can exploit these accurate depth cues, leading to better accuracy in these fields.  
 This work will focus on self-driving cars, while using sparse LiDAR and monocular RGB images. Here, it is desirable to accurately detect and differentiate objects close as well as far away. The LiDAR generates a point cloud of its surroundings, but the limited amount of scan lines results in a high sparsity. LiDARs with 64 scan lines are common and still expensive. The sparse and irregular spaced input points make this task stand out from others. Since a vast amount of applications use LiDAR with a limited amount of scan lines, the industrial relevance is indisputable, currently leading to a very active research domain.
 The reason why this task is challenging is threefold. Firstly, the input is randomly spaced which makes the usage of straightforward convolutions difficult. Secondly, the combination of multiple modalities is still an active area of research, since multiple combinations of sensor fusion are possible, namely early and/or late fusion. This paper will focus on the fusion between RGB info and the LiDAR points. Thirdly, the used annotations are only partially completed. The construction of the pixel-wise ground truth annotations is expensive after all. Our method needs to cope with this constraint. 
 
 The contributions of this paper are:
\begin{itemize}
  \item[(1)] Global and local information are combined in order to accurately complete and correct the sparse input. Monocular RGB images can be used as guidance for this depth completion task. 
  \item[(2)] Confidence maps are learned for both the global and the local branch in an unsupervised manner. The predicted depth maps are weighted by their respective confidence map. This late fusion approach is a fundamental part of the framework.
   \item[(3)] This method ranks first on the KITTI depth completion benchmark with and without using RGB images. Furthermore, it does not require any additional data or postprocessing.
\end{itemize}

The structure of the manuscript will be as follows. Section \ref{related_work} mentions similar prior works regarding depth completion and focuses on the existing challenges. This is followed by a detailed description of our method in section \ref{method}. We further evaluate this method on the popular KITTI dataset in section \ref{experiments}. To conclude, section \ref{conclusion} will wrap up our paper.

\section{Related Work} \label{related_work}
Related works with regards to the depth completion task will be discussed. Attention will be given towards the handling of sparse data and the guidance of LiDAR with other modalities, in particular RGB images.
\subsection{Handling sparse data} 
Completing missing information while also correcting the input has a wide range of applications. Inpainting, denoising and superresolution can all be considered parts of the depth completion task, making depth completion relevant for those specific sub-tasks.

Older methods use handcrafted approaches in order to perform the local upsampling of the sparse input, by usage of complex interpolation techniques. Even more recently, J.~Ku~\etal~\cite{ip_basic} have achieved impressive results without making use of convolutional neural networks (CNNs). They artificially make the input denser by morphological image processing techniques and predict the final depth from this intermediate state. These methods are however prone too errors in the LiDAR frame, making a CNN a more powerful tool for the depth completion task. It's important to know that the 3D LiDAR points are mapped to the 2D plane, making standard 2D convolution a viable option. Despite the more dense input, convolution operations are not designed to operate on this data, since only valid points ought to be considered by the network.

In fact, recent works have also shown that convolutional neural networks can achieve exciting results for this task. Jaritz~\etal~\cite{jaritz} and F.~Ma~\etal~\cite{sparsetodense} both use a deep neural net, while encoding the sparse values with zeros. They argue that a deep network is necessary for this job. We argue that a combination of a local and global network is a more elegant and an intuitive solution, furthermore yielding better results.

Uhrig~\etal~\cite{sparsity} propose sparsity invariant convolutions in order to take into account the sparse input. They perform normalized convolution operations by propagating the validity mask through each layer of their network with maxpooling. Now, the network can be invariant towards the degree of sparsity. Eldesokey~\etal~\cite{eld1} propose a similar solution to take into account the sparsity. Here, a confidence mask is propagated which requires a second convolution for every layer in order to perform the normalization and to generate a the confidence mask for the next layer. We also experiment with uncertainty, but on a higher level, in order to efficiently combine the feature maps extracted by the global and local network. HMS-Net~\cite{hms} goes even further by adopting a multi-scale network and proposing new operations for concatenating, bilinear up-sampling and adding sparse input maps. We find that those operations are not necessary if the sparsity is constant in every frame, since we notice no accuracy gains when including these modified operations. We therefore stick to the conventional convolutions in our method and show that our framework can handle sparse LiDAR data. Furthermore, adding a validity mask to the sparse input shows no effect on the output accuracy which is in line with the findings of Jaritz~\etal~\cite{jaritz}.

\subsection{Guided depth completion}
By now, multiple methods already include RGB data in order to generate better results. How to combine different sensors is still an open research question. Recent works include fusion techniques in order to generate richer features and a better prior for the depth completion task. RGB data will be used to guide our local network. We now discuss recent guidance and fusion techniques.

\subsubsection{Guidance}
In one line of work, Schneider~\etal~\cite{schneider} include RGB information in order to generate sharp edges for the depth predictions. They use pixel-wise semantic annotations to differentiate multiple objects and they use a geodesic distance measure to enforce sharp boundary edges. F.~Ma~\etal~\cite{sparsetodense, self_sup} make use of a ResNet-based deep neural network which takes the 4D, RGB-D, as input. Furthermore, F.~Ma~\etal~\cite{self_sup} take a self-supervised approach which requires temporal data. They now make use of two streams in order to combine the LiDAR data and RGB images in the same feature space, leading to better results. Instead of completing the input immediately Zhang~\etal~\cite{zhang} predict surface normals by leveraging RGB data, leading to a better prior for depth completion. They finally combine these predictions with the sparse depth input to generate the complete depth maps. Like us, they found that completing sparse data from standalone sparse depth samples is a difficult task, proving the importance of RGB guidance.
\newline
\subsubsection{Fusion}
The fusion of multimodal sensor data is not straightforward.
For example Li~\etal~\cite{li} upsample low resolution depth maps guided by the RGB images and take a late fusion approach. In fact, different fusion techniques can be considered: early fusion, late fusion or multi-level fusion. Valada~\etal~\cite{valada} adopt the latter technique by extracting and combining feature maps at different stages in the encoder from multiple input streams. In general most works, such as~\cite{jaritz, eld2}, show that late fusion can achieve better performance. We propose a combination of early and late fusion showing good results on the KITTI benchmark~\cite{kitti}. In our work, early fusion takes the form of a guidance map for our local network extracted from global information. Uncertainty is adopted in the depth predictions to accomplish late fusion. Further, conventional fusion techniques such as adding, concatenating or multiplying feature maps are utilized. 

\begin{figure*}[t]
\noindent
  \begin{center}
    \includegraphics[height=80mm]{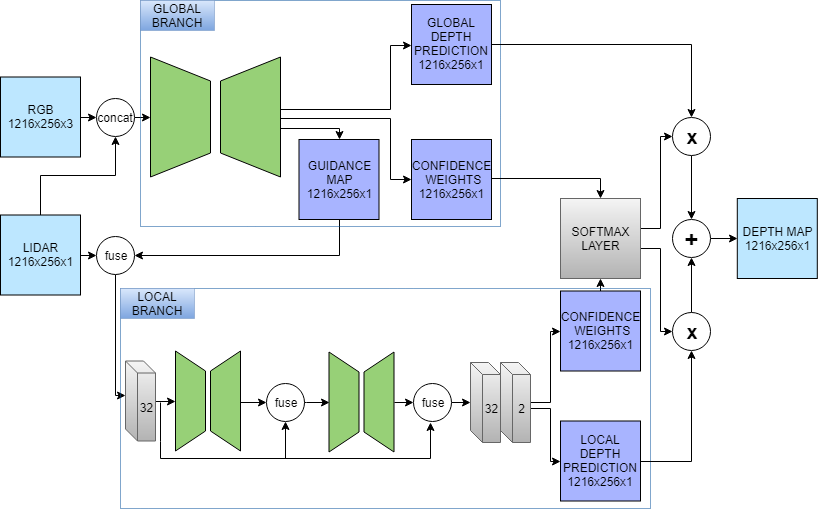}
  \end{center}
  \caption{
    The framework consists of two parts: the global branch on top and the local branch below. The global path outputs three maps: a guidance map, global depth map and a confidence map. The local map predicts a confidence map and a local map by also taking into account the guidance map of the global network. The framework fuses global and local information based on the confidence maps in a late fusion approach. Figure \ref{example} shows that this structure can correct mistakes in the LiDAR input.
  }
  \label{net}
\end{figure*}
\section{Method} \label{method}
Our method acts on a projection of a 3D point cloud to a 2D plane.
Here, the depth completion problem is approached as a regression problem.  Our approach requires supervision by using the ground truth to train our CNN and encodes the missing LiDAR input values with zeros. The targets are reliably composed by using semi-global matching (SGM) and temporal information~\cite{sparsity}, but they are still semi-sparse (around 30\% is filled). Using the sparse input and the semi-sparse ground truth, the convolutional framework makes use of global guidance information to correct artifacts and to upsample the input properly. This correction of artifacts is not explicitly addressed in previous works.

Hence, our method makes use of global and local information in order to complete the input. Since LiDAR is characterized by mistakes due to moving objects and the moving LiDAR itself, both parts are necessary in order get accurate predictions. The local network will interpret local information, whereas the global network extracts global information based on the LiDAR and RGB information. Fusion between the two networks results in a final depth map. We will later show that depth completion does not require a deep network. First, the two parts of the framework will be explained in more detail.
\subsection{Extracting local and global information}
The global branch can be considered as a prior, namely to regularize the features extracted by the local path. Since there are mistakes in the LiDAR input frames, the global information helps the local network to detect these artifacts and reconstruct the sparse input more accurately. We speculate that that the global information is relevant. Firstly, the global network is able to detect (moving) objects and is able to detect structures in the frame that have likely the same depth. Secondly, we expect that a more gradual depth map will be computed in order to prevent sudden and wrong variations in the LiDAR input. This information can be determined by examining the RGB input since borders of objects can be extracted more easily due to its color information. Hence, semantically meaningful information can be extracted.

The local network examines the input LiDAR frame and performs the local up-sampling. To remedy the noisy LiDAR data, we fuse the LiDAR map together with the global guidance map. On the one hand, the reasoning behind this guidance technique is that the local network can further focus on the correct and confident LiDAR points. On the other hand, the global network can reason about objects, its edges and larger structures in the frame. Finally  a residual learning approach has been used in order to keep improving the predictions, implemented by skip connections over the small local networks.

\subsection{Exploiting uncertainty}
We make use of uncertainty in both the global and the local network. Both parts of the framework predict a confidence map. In this way the confidence map acts like a weight map for the final fusion between the two input types. Thus, the weighing is performed per pixel and completely learned by the network in an unsupervised manner. Using this technique, uncertainty in the different network paths is utilized to give more attention to a certain input type, based on the learned confidence weights. The network learns to prefer global information over local information in certain regions. In fact, in locations with accurate and sufficient LiDAR points, the local network will produce depth predictions with a high confidence, whereas global information will be utilized where the LiDAR data is incorrect or scarce, such as at the boundaries of objects. This fusion method is an effective way of combining multiple sensors which is supported by our results in section~\ref{experiments}.

\subsection{Network}
The global network is an encoder-decoder network based on ERFNet~\cite{erfnet} while the local network is a stacked hourglass network. The latter consists of two hourglass modules in order to learn a residual on the original depth predictions, inspired by ResNet~\cite{resnet} and body pose estimation architectures~\cite{hourglass}, with merely 350k parameters in total. Each consists of six layers, has a small receptive field and downsamples only two times by using strided convolutions. No batch normalization~\cite{bn} is present in the first convolution layer and in the encoder of the first hourglass module, since the amount of zeros will skew the layer's parameters, especially when the input sparsity is not constant. The structure of the hourglass module can be found in table~\ref{table_hour}. An ERFNet-based global network has been chosen since it achieves a high accuracy on the Cityscapes' benchmark~\cite{citiscapes} while still being real-time.

The global guidance map is fused with the sparse LiDAR frame, in order to exploit the global info. This resembles early fusion as a guidance for the local network. On the one hand, the global networks provides three output maps: a guidance map with global information, a depth map and a confidence map. On the other hand, the local network provides a depth map and a confidence map. By multiplying the confidence map with its depth map and adding the predictions from both networks, the final prediction is produced. The probability values for the confidence maps are calculated by utilization of the \textit{softmax} function. This selection procedure allows the framework to choose pixels from the global depth map or the adjusted depth values from the stacked hourglass module. Thus, the final depth prediction \(\hat{d}\) exploits the confidence maps X and Y which equates to expression \ref{sofmax_com}. A visualization of the total framework can be found in figure~\ref{net}. 
\begin{equation}
\hat{d}_{out}(i,j) = \frac{e^{X(i,j)}\cdot \hat{d}_{global}(i, j) + e^{Y(i, j)}\cdot \hat{d}_{local}(i, j) }{e^{X(i, j)} + e^{Y(i, j)}}
\label{sofmax_com}
\end{equation} 

\begin{table}[t]
  \caption{Hourglass network.}
  \begin{center}
    \begin{tabular}{c | c c c}
      \hline
      \hline
      \makebox[20mm]{Layer} & \makebox[25mm]{Kernel Size/stride} & 
      \makebox[10mm]{Filters}\\
      \hline
      Conv/Relu & 3x3/2 & 32   \\
      Conv/Relu & 3x3/1 & 64 \\
      Conv/Relu & 3x3/2 & 64  \\
      Conv/Relu & 3x3/1 & 64  \\
      TransConv/BN/Relu & 2x2/2 & 64  \\
      TransConv/BN/Relu & 2x2/2 & 32   \\
      \hline
      \hline
    \end{tabular}
    \label{table_hour}
  \end{center}
\end{table}

\section{Experiments}\label{experiments}
For the experiments a Tesla V100 GPU was used and the code is implemented in Pytorch.
We evaluate our framework by computing the loss on all pixels of the ground truth since not all input pixels of the LiDAR are correct. The KITTI depth completion benchmark~\cite{kitti} is our main focus, since it resembles real-life situations accurately. The KITTI dataset~\cite{dataset} provides 85898 frames for training, 1000 frames for evaluation and 1000 frames for testing.  An ablation study is shown first, followed by a comparison with current state-of-the-art. 

\subsection{Ablation study and analysis}
In all cases we perform data augmentation by flipping the images vertically. Rotating and scaling the LiDAR input while resizing the RGB input had no effect on the final results due to the magnitude of KITTI's dataset. Furthermore, since the LiDAR frame does not provide any information at the top, we crop the inputs to a 1216x256 aspect ratio. We first train both parts of the framework individually and use a pretrained ERFNet on Cityscapes~\cite{citiscapes} for our global network. Afterwards, guidance for the local network is added. Hence, the framework is trained end-to-end and forced to combine the predictions of the two networks based on their certainties with this late fusion approach. We adopt the Adam optimizer~\cite{adam} with learning rate of $10^{-3}$.  

Multiple loss functions were implemented. Our proposed focal-MSE loss, inspired by~\cite{focal}, performed slightly better than the vanilla-MSE loss (by a few mm's) and also better than the popular BerHu loss~\cite{berhu} for the depth prediction task. It is shown in equation \ref{focal}. A focal term has been added in order to give wrongly predicted points during training a slightly higher weight in the loss expression. Furthermore, this regression loss is worth to try in other domains. The loss measures the correctness of the final depth map, the global -and local depth map as shown in equation~\ref{loss_tot}. The weights $w_1, w_2$ are both equal to 0.1 while $w_3$ is equal to 1.
\begin{equation}
\lambda(\hat{y},y) = \frac{1}{n}\sum\limits_{i=1}^n (1 + 0.05\cdot epoch\cdot |y_i-\hat{y_i}|)\cdot (y_i-\hat{y_i})^2
\label{focal}
\end{equation}
\begin{equation}
\Lambda = w_1 \cdot \lambda(\hat{y}_{global}, y) + w_2 \cdot \lambda(\hat{y}_{local}, y) + w_3 \cdot \lambda(\hat{y}_{out}, y)
\label{loss_tot}
\end{equation}

Both the RMSE (root mean squared error) and the MAE (mean absolute error) are used to evaluate on the KITTI benchmark, but we mainly focus on the RMSE since it is the leading metric on the benchmark. The ablation study in table \ref{abl} shows that the combination of a global and a local network leads to impressive results. In fact, our late fusion method, based on uncertainty, contributes to a large accuracy gain. By exploiting the guidance map, we eventually outperform previous works. We furthermore stick to 2 hourglass modules so that the inference time does not increase unnecessarily. Adding batch normalization (BN) to all the convolutions in the local network increases the MAE slightly, due to the high degree of sparsity. We conclude that the local network alone can already achieve good results with only 350k parameters. However, in order to correct mistakes we exploit the global network by predicting uncertainty maps and a guidance map.
\begin{table}[t]
  \caption{Ablation study on KITTI's selected validation set.}
  \begin{center}
    \begin{tabular}{l | c c}
      \hline
      \hline
      \makebox[20mm]{\textbf{ Configuration}} & \makebox[15mm]{RMSE [mm]} & 
      \makebox[15mm]{MAE [mm]} \\
      \hline
      Local Net (LiDAR)            & 995       & 268    \\
      Global Net (RGB)             & 3223      & 1473    \\ 
      Global Net (LiDAR)           & 1020      & 300    \\ 
      Global Net (RGB $\|$ LiDAR)  & 881       & 235    \\ 
      Local+Global+Uncertainty     & 810       & 224    \\
      \; +Guidance skip                 & \textbf{802}       & \textbf{214}    \\
      \; +BN                       & 819       & 223    \\
      \; +Extra Hourglass          & 811       & 222    \\
     
      \hline
      \hline
    \end{tabular}
    \label{abl}
  \end{center}
\end{table}

Table \ref{testset} reports the results on the KITTI testset. We outperform F.~Ma~\etal~\cite{self_sup} (currently ranked first on the KITTI depth completion benchmark~\cite{kitti}) by a significant amount on all metrics, while the frame rate is 4 times higher. Furthermore, we also rank first on the benchmark when we only use LiDAR information in our framework (no RGB images are used) in table \ref{testset}. From this testset data we conclude that the framework can extract semantically meaningful information in order to guide the local network.

\begin{figure}[H]
\noindent
  \begin{center}
    \includegraphics[height=150mm]{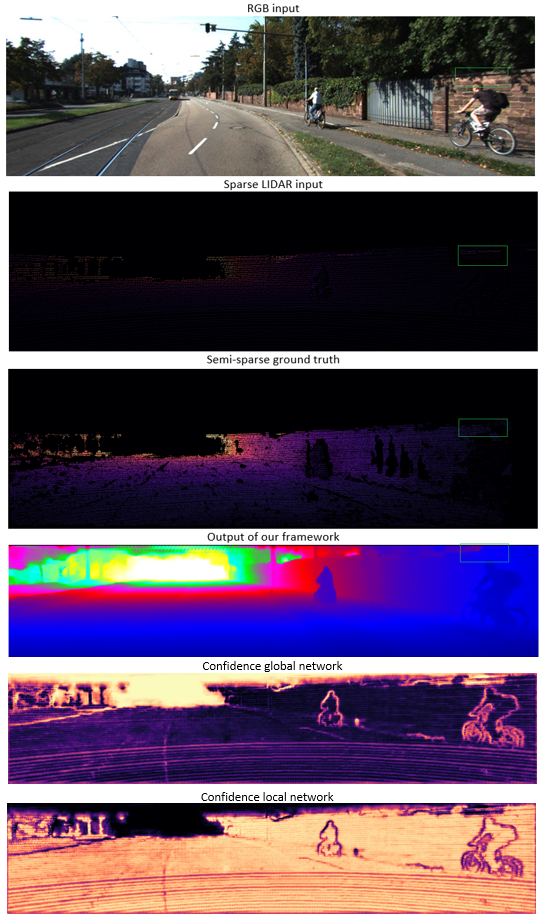}
  \end{center}
  \caption{
    Example on the validation set. The green box shows that our framework successfully corrects the mistakes in the sparse LiDAR input frame.
  }
  \label{example}
\end{figure}

\begin{figure*}[t]
\noindent
  \begin{center}
    \includegraphics[height=75mm]{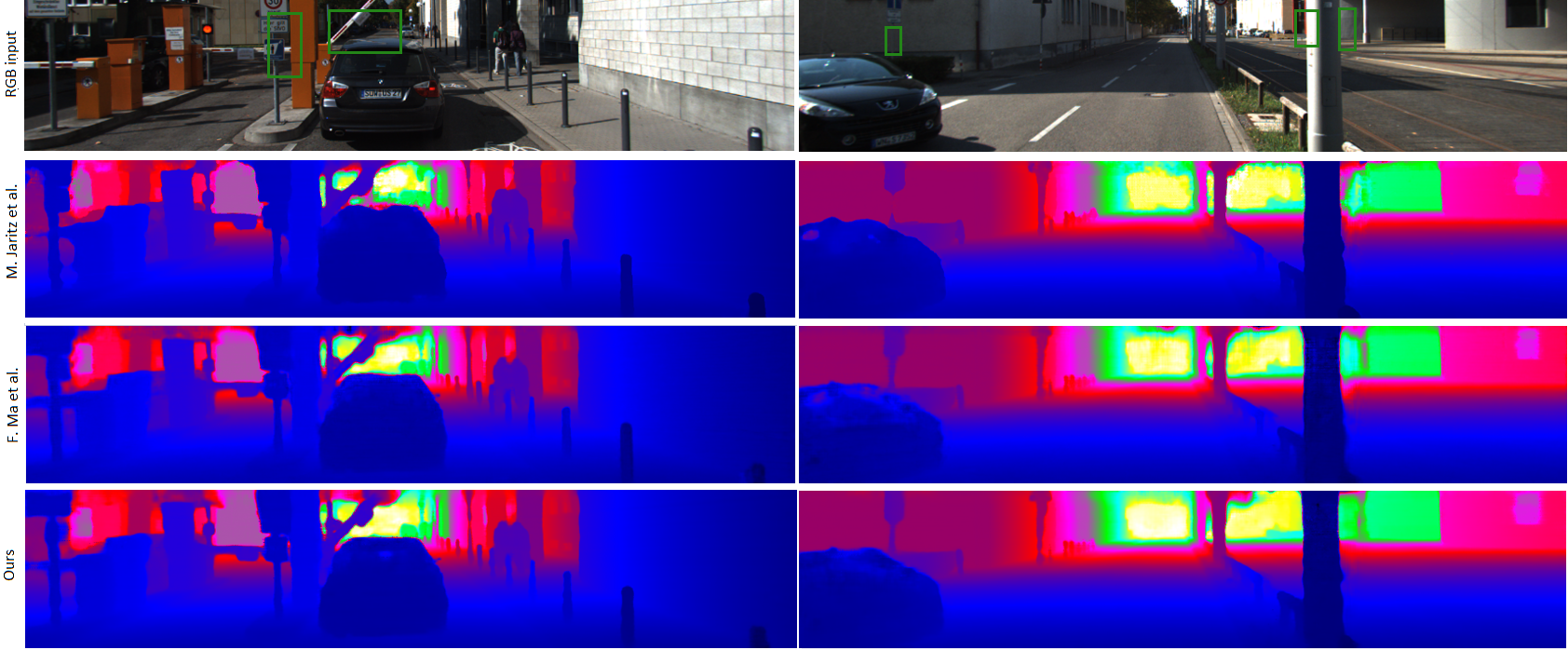}
  \end{center}
  \caption{
    Visual comparison with state-of-the-art. The green box shows the area to focus on in the depth maps. Our method shows better results around objects. For example, on the right of the pillar, the other two methods produce incorrect depth values.
  }
  \label{comparison}
\end{figure*}

Figure \ref{example} displays an example from the validation set. Here, the confidence maps clearly show that the global network is more certain around edges and locations where the LiDAR sensor is incorrect (green box). This proves the effectiveness of this framework. Figure \ref{comparison} demonstrates the differences between our method and other state-of-the-art methods. It shows that we predict more accurate depth values around close as well as far away objects.

\balance

\begin{table}[t]
  \caption{Comparison with state-of-the-art on the testset based on RMSE[mm], MAE[mm] and t[s].}
  \begin{center}
    \begin{tabular}{l | c c c c}
      \hline
      \hline
      \makebox[15mm]{Network} & 
      \makebox[5mm]{RGB} &
      \makebox[10mm]{RMSE} & 
      \makebox[10mm]{MAE} &
      \makebox[10mm]{t} \\
      
      \hline
      SparseConvs~\cite{sparsity} & \xmark & 1601     & 481 & 0.01   \\
      NConv-CNN~\cite{eld2} & \xmark & 1268     & 360 & 0.01   \\
      Spade-sD~\cite{jaritz}  & \xmark   & 1035     & 248 & 0.04   \\
      Sparse-to-Dense~\cite{self_sup} & \xmark   & 954     & 288 & 0.04    \\
      HMS-Net~\cite{hms} & \xmark     & 937     & 258 & 0.02   \\ 
      \textbf{FusionNet (Ours)} & \xmark  & \textbf{923}    & 249 & 0.02\\
      
      \hline
      Spade-RGBsD~\cite{jaritz} & \checkmark     & 918     & 235 & 0.07\\
      NConv-CNN-L1~\cite{eld2} & \checkmark & 859     & 208   & 0.02\\
      HMS-Net\_v2~\cite{hms} & \checkmark    & 842     & 253 & 0.02   \\
      NConv-CNN-L2~\cite{eld2} & \checkmark & 830     & 233   & 0.02   \\
      Sparse-to-Dense~\cite{self_sup} & \checkmark  & 815     & 250 & 0.08    \\ 
      \hline
      \textbf{FusionNet (Ours)} & \checkmark  & \textbf{773}    & 215 & 0.02  \\      \hline
      \hline
    \end{tabular}
    \label{testset}
  \end{center}
\end{table}

\section{Conclusion}\label{conclusion}
We proposed a framework guided by RGB images in order to complete and correct sparse LiDAR frames. The core of the idea is leveraging global information by using a global network. Furthermore, we exploit confidence maps in order to combine both inputs based on the uncertainty in a late fusion approach. We successfully regress towards the semi-sparse ground truth annotations using our focal loss. This method takes 20 ms inference time, hence it meets the real-time requirements for self-driving cars with a large margin. Finally, we evaluated our method on the KITTI dataset where we rank first on the depth completion benchmark. \\
\textbf{Acknowledgement:}
The work was supported by Toyota, and
was  carried  out  at  the  TRACE  Lab  at  KU  Leuven  (Toyota
Research on Automated Cars in Europe - Leuven)

\balance

\end{document}